\documentclass{article}

\usepackage{microtype}
\usepackage{graphicx}
\usepackage{subfigure}
\usepackage{booktabs} 

\usepackage{hyperref}

\usepackage[accepted]{icml2023}

\usepackage{amsmath}
\usepackage{amssymb}
\usepackage{mathtools}
\usepackage{amsthm}
\usepackage{bm}
\usepackage{dsfont}
\usepackage{listings}
\usepackage{tabularx}
\usepackage[capitalize,noabbrev]{cleveref}

\definecolor{mygray}{RGB}{245,245,245}
\definecolor{mygreen}{rgb}{0,0.6,0}
\definecolor{mymauve}{rgb}{0.58,0,0.82}

\theoremstyle{plain}

\theoremstyle{definition}

\theoremstyle{remark}

\usepackage[textsize=tiny]{todonotes}

\usepackage{multirow}
\icmltitlerunning{Random Teachers are Good Teachers}

\begin{document}

\twocolumn[
\icmltitle{Random Teachers are Good Teachers}



\icmlsetsymbol{equal}{*}

\begin{icmlauthorlist}
\icmlauthor{Felix Sarnthein}{eth}
\icmlauthor{Gregor Bachmann}{eth}
\icmlauthor{Sotiris Anagnostidis}{eth}
\icmlauthor{Thomas Hofmann}{eth}            
\end{icmlauthorlist}

\icmlaffiliation{eth}{Department of Computer Science, ETH Zürich, Switzerland}

\icmlcorrespondingauthor{Felix Sarnthein}{safelix@ethz.ch}

\icmlkeywords{Machine Learning, ICML}

\vskip 0.3in
]



\printAffiliationsAndNotice{}  

\begin{abstract}
In this work, we investigate the implicit regularization induced by teacher-student learning dynamics in self-distillation. To isolate its effect, we describe a simple experiment where we consider teachers at random initialization instead of trained teachers. Surprisingly, when distilling a student into such a random teacher, we observe that the resulting model and its representations already possess very interesting characteristics; (1) we observe a strong improvement of the distilled student over its teacher in terms of probing accuracy. (2) The learned representations are data-dependent and transferable between different tasks but deteriorate strongly if trained on random inputs. (3) The student checkpoint contains sparse subnetworks, so-called lottery tickets, and lies on the border of linear basins in the supervised loss landscape. These observations have interesting consequences for several important areas in machine learning: (1) Self-distillation can work solely based on the implicit regularization present in the gradient dynamics without relying on any \textit{dark knowledge}, (2) self-supervised learning can learn features even in the absence of data augmentation, and (3) training dynamics during the early phase of supervised training do not necessarily require label information. Finally, we shed light on an intriguing local property of the loss landscape: the process of feature learning is strongly amplified if the student is initialized closely to the teacher. These results raise interesting questions about the nature of the landscape that have remained unexplored so far. Code is available at \url{www.github.com/safelix/dinopl}.
\end{abstract}

\section{Introduction}
\begin{figure}
    \centering
    \includegraphics[width=0.99\linewidth]{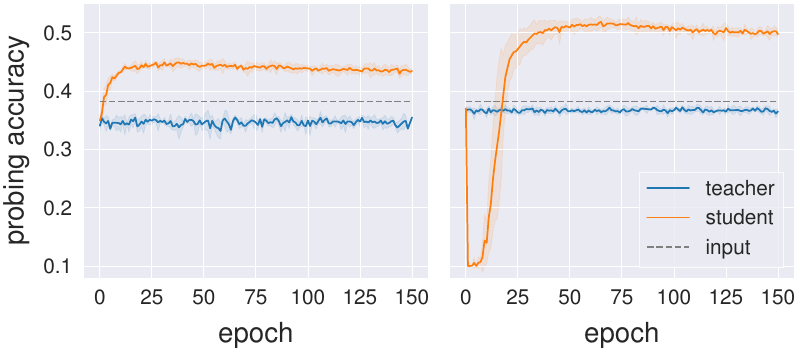}
    \caption{Linear probing accuracies of representations generated by teachers, students, and the flattened input images on \textit{CIFAR10} as a function of training time. \textbf{Left:} \textit{ResNet18}. \textbf{Right:} \textit{VGG11} without batch normalization.}
    \label{fig:init_distill}
    \vspace{-4mm}
\end{figure}

The teacher-student setting is a key ingredient in several areas of machine learning. Knowledge distillation is a common strategy to achieve strong model compression by training a smaller student on the outputs of a larger teacher model, leading to better performance compared to training the small model on the original data only \citep{Bucila2006ModelC, Ba2013DoDN, Hinton2015DistillingNetwork, polino2018model, beyer2022knowledge}. In the special case of self-distillation, where the two architectures match, it is often observed in practice that the student manages to outperform its teacher \citep{Yim_2017_CVPR, Furlanello2018BornNetworks, Yang2018SnapshotDT}. The predominant hypothesis in the literature attests this surprising gain in performance to the so-called \textit{dark knowledge} of the teacher, i.e., its logits encode additional information about the data distribution \citep{Hinton2015DistillingNetwork, Wang2021EmbracingDistillation, xu2018interpreting}.  \\[1mm]
Another area  relying on a teacher-student setup is self-supervised learning where the goal is to learn informative representations in the absence of targets~\citep{Caron2021EmergingTransformers, Grill2020BootstrapLearning, Chen2021ExploringLearning, Zbontar2021BarlowReduction, Assran2022MaskedLearning}. Here, the two models typically receive two different augmentations of a sample, and the student is forced to mimic the teacher's behavior. Such a learning strategy encourages representations that remain invariant to the employed augmentation pipeline, which in turn leads to better downstream performance.  \\[1mm]
Despite its importance as a building block, the teacher-student setting itself remains very difficult to analyze as its contribution is often overshadowed by stronger components in the pipeline, such as \textit{dark knowledge} in the trained teacher or the inductive bias of data augmentation. In this work, we take a step towards simplifying and isolating the key components in the setup by devising a very simple experiment; instead of working with a trained teacher, we consider teachers at random initialization, stripping them from any data dependence and thus removing any \textit{dark knowledge}. We also remove augmentations, making the setting completely symmetric between student and teacher and further reducing inductive bias. Counter-intuitively, we observe that even in this setting, the student still manages to learn from its teacher and even exceed it significantly in terms of representational quality, measured through linearly probing the features (see Fig.~\ref{fig:init_distill}). This result shows the following: (1) Even in the absence of \textit{dark knowledge}, relevant feature learning can happen for the student in the setting of self-distillation. (2) Data augmentation is the main but not only ingredient in non-contrastive self-supervised learning that leads to representation learning. \\[1mm]
Surprisingly, we find that initializing the student close to the teacher further amplifies the implicit regularization present in the dynamics. This is in line with common practices in non-contrastive learning, where teacher and student are usually initialized closely together and only separated through small asymmetries in architecture and training protocol \citep{Grill2020BootstrapLearning, Caron2021EmergingTransformers}. We study this locality effect of the landscape and connect it with the \textit{asymmetric valley} phenomenon observed in \citet{He2019AsymmetricMinima}. \\[1mm]
The improvement in probing accuracy suggests that some information about the data is incorporated into the network's weights.
To understand how this information is retained, we compare the behavior of supervised optimization to finetuning student networks. 
We find that some of the learning dynamics observable during the early phase of supervised training also occur during random teacher distillation.
In particular, the student already contains sparse subnetworks and reaches the border of linear basins in the supervised loss landscape. This contrasts \citep{Frankle2020TheTraining} where training on a concrete learning task for a few epochs is essential. 
Ultimately, these results suggest that label-independent optimization dynamics exist and allow exploring the supervised loss landscape to a certain degree. 

\section{Related Work}
Several works in the literature aim to analyze self-distillation and its impact on the student.  \citet{Phuong2019TowardsUK} prove a generalization bound that establishes fast decay of the risk in the case of linear models. \citet{Mobahi2020Self-DistillationSpace} demonstrate an increasing regularization effect through repeated distillation for kernel regression. \citet{Guangda2020KDNTK} consider a similar approach and rely on the fact that very wide networks behave very similarly to the neural tangent kernel \citep{Jacot2018NTK} and leverage this connection to establish risk bounds. \citet{Allen-Zhu2020TowardsLearning} on the other hand, study more realistic width networks and show that if the data satisfies a certain multi-view property, ensembling and distilling is provably beneficial. \citet{Yuan2020labelsmoothing} study a similar setup as our work by considering teachers that are not perfectly pre-trained but of weaker (but still far from random) nature. They show that the \textit{dark knowledge} is more of a regularization effect and that a similar boost in performance can be achieved by label smoothing. \citet{Stanton2021DoesWork} further question the relevance of \textit{dark knowledge} by showing that students outperform their teacher without fitting the \textit{dark knowledge}. We would like to point out however that we study completely random teachers and our loss function does not provide the hard labels for supervisory signal, making our task completely independent of the targets. \\[1mm]
Self-supervised learning can be broadly split into two categories, contrastive and non-contrastive methods. Contrastive methods rely on the notion of negative examples, where features are actively encouraged to be dissimilar if they stem from different examples \citep{Chen2020ARepresentations, Schroff2015FaceNet, Oord2018RepresentationLW}. Non-contrastive methods follow our setting more closely as only the notion of positive examples is employed \citep{Caron2021EmergingTransformers, Grill2020BootstrapLearning, Chen2021ExploringLearning}. While these methods enjoy great empirical successes, a theoretical understanding is still largely missing. \citet{Tian2021UnderstandingPairs} investigate the collapse phenomenon in non-contrastive learning and show in a simplified setting how the stop gradient operation can prevent it. \citet{wang2022towards} extend this work and prove in the linear setting how a data-dependent projection matrix is learned. \citet{Zhang2022HowLearning} explore a similar approach and prove that \textit{SimSiam} \citep{Chen2021ExploringLearning} avoids collapse through the notion of extra-gradients. \citet{Anagnostidis2022TheMemorization} show that strong representation learning occurs with heavy data augmentations even if random labels are used. Despite this progress on the optimization side, a good understanding of feature learning has largely remained elusive.  \\[1mm]
The high-dimensional loss landscapes of neural networks remain very mysterious, and their properties play a crucial role in our work. \citet{Safran2017SpuriousNetworks} prove that spurious local minima exist in the teacher-student loss of two-layer ReLU networks. \citet{Garipov2018LossDNNs, Draxler2018EssentiallyLandscape} show that two SGD solutions are always connected through a non-linear valley of low loss. \citet{Frankle2018TheNetworks, Frankle2019LinearHypothesis, Frankle2020TheTraining} investigate the capacity of over-parameterized networks through pruning of weights. They find that sparse sub-networks develop already very early in neural network training. \citet{Zaidi2022WhenWork, Benzing2022RandomThem} investigate random initializations in supervised loss landscapes. Still, the field lacks a convincing explanation as to how simple first-order gradient-based methods such as SGD manage to navigate the landscape so efficiently.

\section{Setting} \label{sec:setting}

\paragraph{Notation.} Let us set up some notation first. We consider a family of parametrized functions $\mathcal{F}=\{f_{\bm{\theta}}:\mathbb{R}^{d} \xrightarrow[]{}\mathbb{R}^{m}\big{|}\bm{\theta} \in\Theta \}$ where $\bm{\theta}$ denotes the (vectorized) parameters of a given model and $\Theta$ refers to the underlying parameter space. In this work, we study the teacher-student setting, i.e., we consider two models $f_{\bm{\theta}_T}$ and $f_{\bm{\theta}_S}$ from the same function space $\mathcal{F}$. We will refer to $f_{\bm{\theta}_T}$ as the teacher model and to $f_{\bm{\theta}_S}$ as the student model. Notice that here we assume that both teacher and student have the same architecture unless otherwise stated. Moreover, assume that we have access to $n \in \mathbb{N}$ input-output pairs $(\bm{x}_1,y_1), \dots, (\bm{x}_n,y_n) \stackrel{i.i.d.}{\sim}\mathcal{D}$ distributed according to some probability measure $\mathcal{D}$, where $\bm{x}_i \in \mathbb{R}^{d}$ and $y_i \in \{0, \dots, K-1\}$ encodes the class membership for one of the $K \in \mathbb{N}$ classes.

\paragraph{Supervised.} The standard learning paradigm in machine learning is supervised learning, where a model $f_{\bm{\theta}} \in \mathcal{F}$ is chosen based on empirical risk minimization, i.e., given a loss function $l$, we train a model to minimize 
$$L(\bm{\theta}) := \sum_{i=1}^{n}l(f_{\bm{\theta}}(\bm{x}_i), y_i).$$
Minimization of the objective is usually achieved by virtue of standard first-order gradient-based methods such as SGD or ADAM \citep{kingma2017adam}, where parameters $\bm{\theta} \sim \text{INIT}$ are randomly initialized and then subsequently updated based on gradient information.

\begin{figure}
    \centering
    \includegraphics[width=0.99\linewidth]{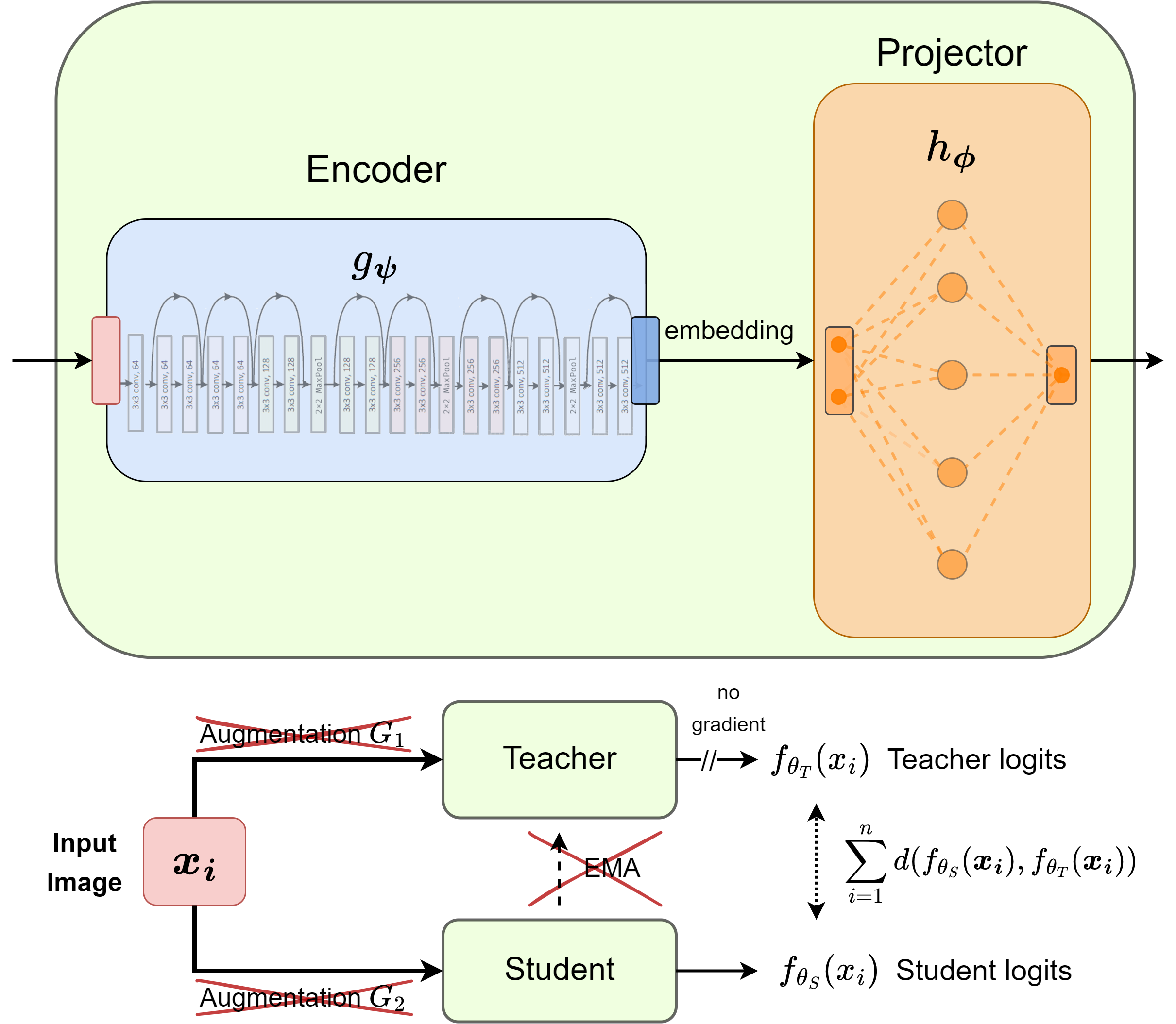}
    \caption{Schematic drawing of the teacher-student setup. The model consists of an encoder and projector. The same image is passed to both student and teacher, and the outputs of the projectors are compared. The student weights are then adjusted to mimic the output of the teacher. In this work, we consider a simplified setting without augmentations and without teacher updates such as EMA.}
    \label{fig:teacher-student}
    \vspace{-2mm}
\end{figure}

\paragraph{Teacher-Student Loss.} A similar but distinct way to perform learning is the teacher-student setting. Here we first fix a teacher model $f_{\bm{\theta}_T}$ where $\bm{\theta}_T$ is usually a parameter configuration arising from training in a supervised fashion on the same task. The task of the student $f_{\bm{\theta}_S}$is then to mimic the teacher's behavior on the training set by  minimizing a distance function $d$ between the two predictions,
\begin{equation}
    L(\bm{\theta}_S) := \sum_{i=1}^{n}d\left(f_{\bm{\theta}_S}(\bm{x}_i),f_{\bm{\theta}_T}(\bm{x}_i)\right).
    \label{eq:distill-loss}
\end{equation}
We have summarized the setting schematically in Fig.~\ref{fig:teacher-student}. We experiment with several choices for the distance function but largely focus on the KL divergence. We remark that the standard definition of distillation \citep{Hinton2015DistillingNetwork} consider a combination of losses of the form 
$$L(\bm{\theta}_S) := \sum_{i=1}^{n}d\left(f_{\bm{\theta}_S}(\bm{x}_i),f_{\bm{\theta}_T}(\bm{x}_i)\right) + \beta \sum_{i=1}^{n}l(f_{\bm{\theta}_S}(\bm{x}_i), y_i),$$
for $\beta >0$, thus the objective is also informed by the true labels $y$. Here we set $\beta=0$ to precisely test how much performance is solely due to the implicit regularization present in the learning dynamics and the inductive bias of the model. \\[1mm]
Somewhat counter-intuitively, it has been observed in many empirical works that the resulting student often outperforms its teacher. It has been hypothesized in many prior works that the teacher logits $f_{\bm{\theta}_T}(\bm{x})$ encode some additional, relevant information for the task that benefits learning (\textit{dark knowledge}), i.e., wrong but similar classes might have a non-zero probability under the teacher model \citep{Hinton2015DistillingNetwork, Wang2021EmbracingDistillation, xu2018interpreting}. In the following, we will explore this hypothesis by systematically destroying the label information in the teacher. 

\paragraph{Non-Contrastive.} Self-supervised learning is a recently developed methodology enabling the pretraining of vision models on large-scale unlabelled image corpora, akin to the autoregressive loss in natural language processing \citep{Devlin2019BERT}. A subset of these approaches is formed by non-contrastive methods. Consider a set of image augmentations $\mathcal{G}$ where any $G \in \mathcal{G}$ is a composition of standard augmentation techniques such as random crop, random flip, color jittering, etc. The goal of non-contrastive learning is to learn a parameter configuration that is invariant to the employed data augmentations while avoiding simply collapsing to a constant function. Most non-contrastive objectives can be summarized to be of the form
$$L(\bm{\theta}_S) := \sum_{i=1}^{n}\mathbb{E}_{G_1,G_2}\left[d\left(f_{\bm{\theta}_S}(G_1(\bm{x}_i)), f_{\bm{\theta}_T}(G_2(\bm{x}_i))\right)\right],$$
where the expectation is taken uniformly over the set of augmentations $\mathcal{G}$. We summarize this pipeline schematically in Fig.~\ref{fig:teacher-student}. While the teacher does not directly receive any gradient information, the parameters $\bm{\theta}_T$ are often updated based on an exponentially weighted moving average,
$$\bm{\theta}_T \longleftarrow (1-\gamma) \bm{\theta}_T + \gamma\bm{\theta}_S$$
which is applied periodically at a fixed frequency. In this work, we will consider a simplified setting without augmentations and where the teacher remains frozen at random initialization, $\gamma = 0$.

\paragraph{Probing.} Since minimizing the teacher-student loss is a form of unsupervised learning if the teacher itself has not seen any labels, we need a way to measure the quality of the resulting features. Here we rely on the idea of probing representations, a very common technique from self-supervised learning \citep{Chen2021ExploringLearning, Chen2020ARepresentations, Caron2021EmergingTransformers, Bardes2021VICReg:Learning, Grill2020BootstrapLearning}. As illustrated in Fig.~\ref{fig:teacher-student}, the network is essentially split into an encoder $g_{\bm{\psi}}:\mathbb{R}^{d} \xrightarrow[]{}\mathbb{R}^r$ and a projector $h_{\bm{\phi}}:\mathbb{R}^{r} \xrightarrow[]{}\mathbb{R}^{m}$ where it holds that $f_{\bm{\theta}} = h_{\bm{\phi}} \circ g_{\bm{\psi}}$. The encoder is usually given by the backbone of a large vision model such as \textit{ResNet}~\citep{he2015resnet} or \textit{VGG}~\citep{Simonyan2014VGG}, while the projector is parametrized by a shallow MLP. We then \textit{probe} the representations $g_{\bm{\psi}}$ by learning a linear layer on top, where we now leverage the label information $y_1, \dots, y_n$. Notice that the weights of the encoder remain frozen while learning the linear layer. The idea is that a linear model does not add more feature learning capacity, and the resulting probing accuracy hence provides an adequate measure of the quality of the representations. Unless otherwise stated, we perform probing on the \textit{CIFAR10} dataset \citep{Krizhevsky2009CIFAR} and aggregate mean and standard deviation over three runs.

\section{Random Teacher Distillation}
\paragraph{Distillation.} Let us denote by $\bm{\theta} \sim \textit{INIT}$ a randomly initialized parameter configuration, according to some standard initialization scheme $\text{INIT}$. Throughout this text, we rely on \textit{Kaiming} initialization \citep{He2015KaimingInit}. In standard self-distillation, the teacher is a parameter configuration ${\bm{\theta}}_T^{(l)}$ resulting from training in a supervised fashion for $l\in\mathbb{N}$ epochs on the task $\{(\bm{x}_i, y_i)\}_{i=1}^{n}$.

In a next step, the teacher is then distilled into a student, i.e., the student is trained to match the outputs of the pre-trained teacher $f_{{\bm{\theta}}_T^{(l)}}$. In this work, we change the nature of the teacher and instead consider a teacher at random initialization $\bm{\theta}_T \sim \text{INIT}$ (we drop the superscript $0$ for convenience). The teacher has thus not seen any data at all and is hence of a similar (bad) quality as the student. This experiment, therefore, serves as the ideal test bed to measure the implicit regularization present in the optimization itself without relying on any \textit{dark knowledge} about the target distribution. Due to the absence of targets, the setup also closely resembles the learning setting of non-contrastive methods. Through that lens, our experiment can also be interpreted as a non-contrastive pipeline without \emph{augmentations} and exponential moving average. \\[1mm]
We minimize the objective \eqref{eq:distill-loss} with the ADAM optimizer \citep{kingma2017adam} using a learning rate $\eta=0.001$. We analyze two encoder types based on the popular \textit{ResNet18} and \textit{VGG11} architectures, and similarly to \citet{Caron2021EmergingTransformers}, we use a $2$-hidden layer MLP with an $L_2$ bottleneck, as a projector. To assess whether batch-dependent statistics play a role, we remove the batch normalization layers \citep{Ioffe2015BatchNorm} from the \textit{VGG11} architecture. For more details on the architecture and hyperparameters, we refer to App.~\ref{app:details}. 
\begin{table}[h]
\begin{center}
\begin{small}
\begin{sc}
\begin{tabular}{lcccc}
\toprule
Dataset & Model & Teacher & Student & Input \\
\midrule
\multirow{ 2}{*}{ \textit{CIFAR10}}  
    & \textit{ResNet18} & $35.50$ & $\mathbf{46.02}$ & \multirow{2}*{$39.02$} \\[1mm]
    & \textit{VGG11} & $36.55$ & $\mathbf{51.98}$ & \\
\midrule
\multirow{ 2}{*}{ \textit{CIFAR100}}  
    & \textit{ResNet18} & $11.58$ & $\mathbf{21.50}$ & \multirow{2}*{$14.07$} \\[1mm]
    & \textit{VGG11} & $12.05$ & $\mathbf{26.62}$ & \\
\midrule
\multirow{ 2}{*}{ \textit{STL10}}  
    & \textit{ResNet18} & $24.24$ & $\mathbf{40.58}$ & \multirow{2}*{$31.51$} \\[1mm]
    & \textit{VGG11} & $24.67$ & $\mathbf{46.20}$ & \\
\midrule
\multirow{ 2}{*}{ \textit{TinyImageNet}} 
    & \textit{ResNet18} & $4.85$ & $\mathbf{10.40}$ & \multirow{2}*{$3.28$} \\[1mm]
    & \textit{VGG11} & $5.25$ & $\mathbf{12.88}$ & \\
\bottomrule
\end{tabular}
\end{sc}
\end{small}
\end{center}
\caption{Linear probing accuracies (in percentage) of the representations for various datasets for teacher, student and flattened input images. Students outperform the baselines in all cases.}
\label{tab:distill}
\end{table}

We display the linear probing accuracy of both student and teacher as a function of training time in Fig.~\ref{fig:init_distill}. We follow the protocol of non-contrastive learning and initialize the student closely to the teacher. We will expand more on this choice of initialization in the next paragraph. Note that while the teacher remains fixed throughout training, accuracies can vary due to stochastic optimization of linear probing. The dashed line represents the linear probing accuracy obtained directly from the (flattened) inputs. We clearly see that the student significantly outperforms its teacher throughout the training. Moreover, it also improves over probing on the raw inputs, demonstrating that not simply less signal is lost due to random initialization but rather that meaningful learning is performed. We expand our experimental setup to more datasets, including \textit{CIFAR100} \citep{Krizhevsky2009CIFAR}, \textit{STL10} \citep{Coates2011STL} and \textit{TinyImageNet} \citep{Le2015TinyImageNet}. We summarize the results in Table~\ref{tab:distill}. We observe that across all tasks, distilling a random teacher into its student proves beneficial in terms of probing accuracy. For further ablations on the projection head, we refer to the App.~\ref{app:projector}. Moreover, we find similar results for more architectures and $k$-NN instead of linear probing in App.~\ref{app:more-stuff}.

\begin{figure}[t]
    \centering
    \includegraphics[width=0.99\linewidth]{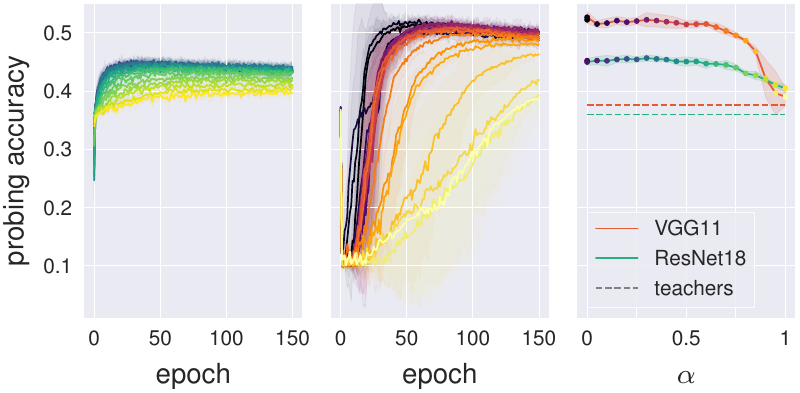}
   \caption{Linear probing accuracies as a function of the locality parameter $\alpha$ on \textit{CIFAR10}. The color gradient (bright $\rightarrow$ dark) reflects the value of $\alpha$ ($0 \rightarrow1 $) for \textit{ResNet18} in green and \textit{VGG11} in red tones.  
    \textbf{Left:\ }\textit{ResNet18}. 
    \textbf{Middle:\ }\textit{VGG11}. 
    \textbf{Right:\ }Summary.
    }
    \label{fig:locality}
    \vspace{-2mm}
\end{figure}

\paragraph{Local Initialization.}
It turns out that the initialization of the student and its proximity to the teacher plays a crucial role. To that end, we consider initializations of the form $$\bm{\theta}_S(\alpha) = \frac{1}{\delta}\left((1-\alpha){\bm{\theta}}_T + \alpha\tilde{\bm{\theta}}\right),$$ where $\tilde{\bm{\theta}} \sim \text{INIT}$ is a fresh initialization, $\alpha \in [0,1]$ and $\delta =\sqrt{\alpha^2 + (1 - \alpha)^2}$ ensures that the variance remains constant $\forall \alpha \in [0,1 ]$. By increasing $\alpha$ from $0$ towards $1$, we can gradually separate the student initialization from the teacher and ultimately reach the more classical setup of self-distillation where the student is initialized independently from the teacher. Note, that in the non-contrastive learning setting, teacher and student are initialized at the same parameter values (i.e., $\alpha=0$), and only minor asymmetries in the architectures lead to different overall functions. \\[1mm]
We now study how the locality parameter $\alpha$ affects the resulting quality of the representations of the student in our setup. In Fig.~\ref{fig:locality}, we display the probing accuracy as a function of the training epoch for different choices of $alpha$. Furthermore, we summarize the resulting accuracy of the student as a function of the locality parameter $\alpha$. Surprisingly, we observe that random teacher distillation behaves very similarly for all $\alpha \in [0, 0.6]$. Increasing $\alpha$ more slows down the convergence and leads to worse overall probing performance. However, even initializing the student independently of the teacher ($\alpha=1$) results in a considerable improvement over the teacher. In other words, we show that representation learning can occur in self-distillation for any random teacher without \textit{dark knowledge}. To the best of our knowledge, we are the first to observe such a locality phenomenon in the teacher-student landscape. We investigate this phenomenon in more detail in the next section and, for now, if not explicitly stated otherwise, use initializations with small locality parameter $\alpha \sim 10^{-10}$. \citet{Safran2017SpuriousNetworks} prove that spurious local minima exist in the teacher-student loss of two-layer ReLU networks. We speculate that this might be the reason why initializing students close to the teacher is beneficial, and provide evidence in App.~\ref{app:metrics}

\begin{figure}[t]
    \centering
    \includegraphics[width=0.99\linewidth]{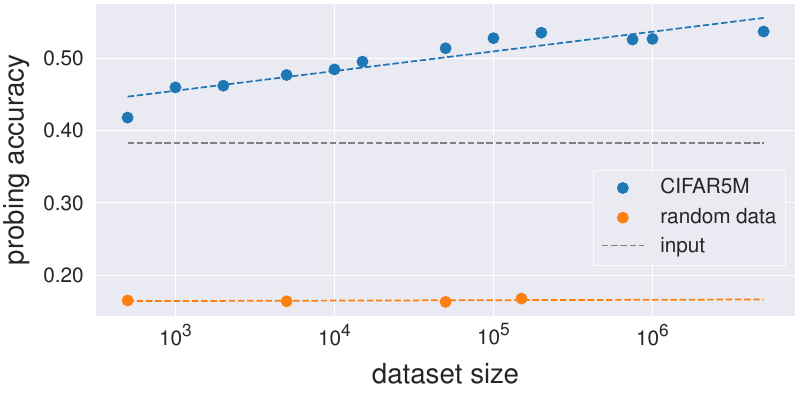}
    \caption{Linear probing accuracies of a \textit{VGG11} trained on \textit{CIFAR5M} or Gaussian noise inputs and evaluated on \textit{CIFAR10} as a function of sample size $n$. Representations are data dependent.}
    \label{fig:dataset_size}
    \vspace{-2mm}
\end{figure}

\paragraph{Data-Dependence.} In a next step, we aim to understand better to which degree the learned features are data dependent, i.e., tuned to the particular input distribution $\bm{x} \sim p_{\bm{x}}$. While the improvement over the raw input probe already suggests non-trivial learning, we want to characterize the role of the input data more precisely. \\[1mm]
As a first experiment, we study how the improvement of the student over the teacher evolves as a function of the sample size $n$ involved in the teacher-student training phase. We use the \textit{CIFAR5M} dataset, where the standard \textit{CIFAR10} dataset has been extended to $5$ million data points using a generative adversarial network \citep{nakkiran2021cifar5m}. We train the student for different sample sizes in the interval $[5\times10^2, 5\times10^6]$ and probe the learned features on the standard \textit{CIFAR10} training and test set. We display the resulting probing accuracy as a function of sample size in Fig.~\ref{fig:dataset_size} (blue line). Indeed, we observe a steady increase in the performance of the student as the size of the data corpus grows, highlighting that data-dependent feature learning is happening. \\[1mm]
As further confirmation, we replace the inputs $\bm{x}_i \sim p_{\bm{x}}$ with pure Gaussian noise, i.e. $\bm{x}_i \sim \mathcal{N}(\bm{0}, \sigma^2\mathds{1})$, effectively removing any relevant structure in the samples. The linear probing, on the other hand, is again performed on the clean data. This way, we can assess whether the teacher-student training is simply moving the initialization in a favorable way (e.g. potentially uncollapsing it), which would still prove beneficial for meaningful tasks. We display the probing accuracy for these random inputs in Fig.~\ref{fig:dataset_size} as well (orange line) and observe that such random input training does not lead to an improvement of the student across all dataset sizes. This is another indication that data-dependent feature learning is happening, where in this case, adapting to the noise inputs of course proves detrimental for the clean probing.

\paragraph{Transferability.} As a final measure for the quality of the learned features, we test how well a set of representations obtained on one task transfers to a related but different task. More precisely, we are given a source task $\mathcal{A}=\{(\bm{x}_i, y_i)\}_{i=1}^{n} \stackrel{i.i.d.}{\sim} \mathcal{D}_{\mathcal{A}}$ and a target task $\mathcal{B}=\{(\bm{x}_i, y_i)\}_{i=1}^{\tilde{n}} \stackrel{i.i.d.}{\sim} \mathcal{D}_{\mathcal{B}}$ and assume that both tasks are related, i.e., some useful features on $\mathcal{A}$ also prove to be useful on task $\mathcal{B}$.  We first use the source task $\mathcal{A}$ to perform random teacher distillation and then use the target task $\mathcal{B}$ to train and evaluate the linear probe. Clearly, we should only see an improvement in the probing accuracy over the (random) teacher if the features learned on the source task encode relevant information for the target task as well. We use \textit{TinyImageNet} as the source task and evaluate on \textit{CIFAR10}, \textit{CIFAR100}, and \textit{STL10} as target tasks for our experiments. We illustrate the results in Table~\ref{tab:distill_transfer} and observe that transfer learning occurs. This suggests that the features learned by random teacher distillations can encode common properties of natural images which are shared across tasks.

\begin{table}[t]
\vskip 0.15in
\begin{center}
\begin{small}
\begin{sc}
\begin{tabular}{lccc}
\toprule
Dataset & Model & Teacher & Student \\
\midrule
\multirow{ 2}{*}{ \textit{CIFAR10}}  
    & \textit{ResNet18} & $35.50$ & $\mathbf{46.06}$  \\[1mm]
    & \textit{VGG11} & $36.55$ & $\mathbf{52.45}$ \\
\midrule
\multirow{ 2}{*}{ \textit{CIFAR100}}  
    & \textit{ResNet18} & $11.58$ & $\mathbf{22.60}$  \\[1mm]
    & \textit{VGG11} & $12.05$ & $\mathbf{27.49}$  \\
\midrule
\multirow{ 2}{*}{ \textit{STL10}}  
    & \textit{ResNet18} & $24.24$ & $\mathbf{41.42}$ \\[1mm]
    & \textit{VGG11} & $24.67$ & $\mathbf{45.86}$  \\
\bottomrule
\end{tabular}
\end{sc}
\end{small}
\end{center}
\vskip -0.11in
\caption{Linear probing accuracies of the representations for various datasets for teacher and student. Students distilled from random teachers on \textit{TinyImageNet} generalize out of distribution.}
\label{tab:distill_transfer}
\end{table}

\section{Loss and Probing Landscapes}
\paragraph{Visualization.} We now revisit the locality property identified in the previous section, where initializations with $\alpha$ closer to zero outperformed other configurations. To gain further insight into the inner workings of this phenomenon, we visualize the teacher-student loss landscape as well as the resulting probing accuracies as a function of the model parameters. Since the loss function is a very high-dimensional function of the parameters, only slices of it can be visualized at once. More precisely, given two directions $\bm{v}_1, \bm{v}_2$ in parameter space, we form a visualization plane of the form
$$\bm{\theta}(\lambda_1, \lambda_2) = \lambda_1 \bm{v}_1 + \lambda_2 \bm{v}_2, \hspace{5mm} (\lambda_1, \lambda_2) \in [0,1]^2$$
and then collect loss and probing values at a certain resolution. Such visualization strategy is very standard in the literature, see e.g., \citet{Li2018VisualizingNets, Garipov2018LossDNNs, Izmailov2021WhatLike}. Denote by $\bm{\theta}^{*}_S(\alpha)$ the student trained until convergence initialized with locality parameter $\alpha$. We study two choices for the landscape slices. First, we refer to a \textit{non-local view} as the plane defined by the random teacher $\bm{\theta}_T$, the student at a fresh initialization $\bm{\theta}_S(1)$ and the resulting trained student ${\bm{\theta}}_S^{*}(1)$, i.e., we set $\bm{v}_1=\bm{\theta}_S(1) - \bm{\theta}_T$ and $\bm{v}_2={\bm{\theta}}_S^{*}(1) - \bm{\theta}_T$.
As a second choice, we refer to a \textit{shared view} as the plane defined by the random teacher $\bm{\theta}_T$, the trained student starting from a fresh initialization ${\bm{\theta}}_S^{*}(1)$ and the trained student ${\bm{\theta}}_S^{*}(0)$ initialized closely to the teacher, i.e., we set $\bm{v}_1={\bm{\theta}}_S^{*}(0) - \bm{\theta}_T$ and $\bm{v}_2={\bm{\theta}}_S^{*}(1) - \bm{\theta}_T$. 
Note that $\alpha$ is not exactly zero but around $10^{-10}$. 

\begin{figure}[t]
    \centering
    \includegraphics[width=0.99\linewidth]{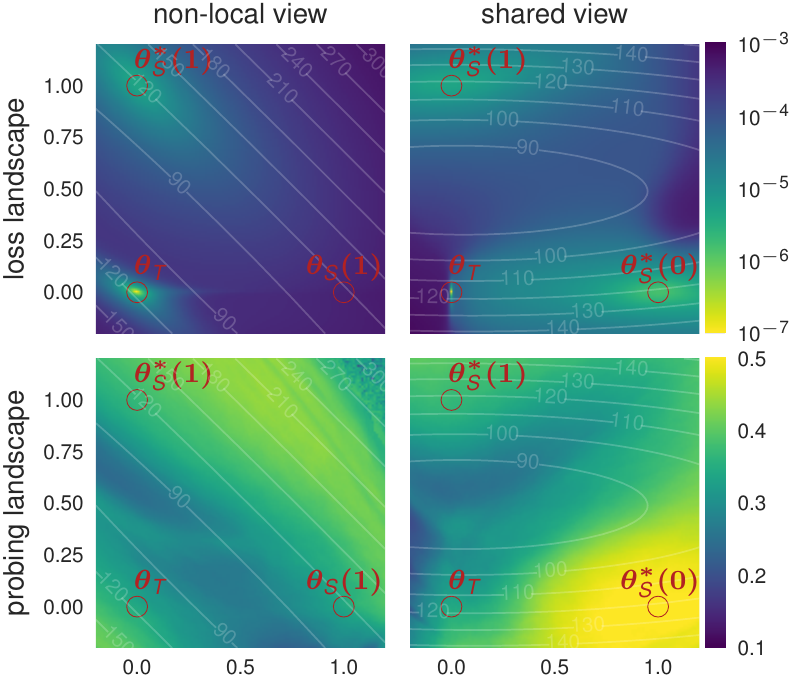}
    \caption{Visualization of the loss and probing landscape.  The left column corresponds to the \textit{non-local view} with $\alpha=1$. The right column depicts the \textit{shared view}, containing both the local ($\alpha=0)$ and the non-local solution ($\alpha=1)$. The first row displays the loss landscape and the second one shows the probing landscape. Contours lines represent $||\bm{\theta}||_2$, orthogonal projections are in App.~\ref{app:more-loss}.}
    \label{fig:losslandscape}
    \vspace{-6mm}
\end{figure}

We show the results in Fig.~\ref{fig:losslandscape}, where the left and right columns represent the \textit{non-local} and the \textit{shared view} respectively, while the first and the second row display loss and probing landscapes respectively.
Let us focus on the  \textit{non-local view} first. Clearly, for $\alpha=1$ the converged student $\bm{\theta}_S^{*}(1)$ ends up in a qualitatively different minimum than the teacher, i.e., the two points are separated by a significant loss barrier. This is expected as the student is initialized far away from the teacher. Further, we see that the probing landscape is largely unaffected by moving from the initialization $\bm{\theta}_S(0)$ to the solution $\bm{\theta}_S^{*}(0)$, confirming our empirical observation in Fig.~\ref{fig:locality} that far way initialized students only improve slightly.
The \textit{shared view} reveals more structure. We see that although it was initialized very closely to the teacher, the student $\bm{\theta}_S^{*}(0)$ moved considerably. While the loss barrier is lower as in the case of $\bm{\theta}^{*}_S(1)$, it is still very apparent that $\bm{\theta}_S^{*}(0)$ settled for a different, local minimum that coincides with a region of high probing accuracy. This is surprising as the teacher itself is the global loss minimum. For more visualizations, including the loss landscape for the encoder, we refer to App.~\ref{app:more-loss}.

\paragraph{Asymmetric valleys.} A striking structure in the loss landscape of the \textit{shared view} is the very pronounced asymmetric valley around the teacher $\bm{\theta}_T$. While there is a very steep increase in loss towards the left of the view (dark blue), the loss increases only gradually in the opposite direction (light turquoise) and quickly decreases into the local minimum of the converged student $\bm{\theta}_S^{*}(0)$. Surprisingly, this direction orthogonal to the cliff identifies a region of high accuracy in the probing landscape. A fact remarkably in line with this situation is proven by \citet{He2019AsymmetricMinima}. They show that being on the flatter side of an asymmetric valley (i.e., towards $\bm{\theta}_S^{*}(0)$) provably leads to better generalization compared to lying in the valley itself (i.e., $\bm{\theta}_T$). Initializing the student closely to the teacher seems to capitalize on that fact and leads to systematically better generalization. Still, it remains unclear why such an asymmetric valley is only encountered close to the teacher and not for initializations with $\alpha=1$. We leave a more in-depth analysis of this phenomenon for future work.

\begin{figure}
    \centering
    \includegraphics[width=0.9\linewidth, trim={0pt, 40pt, 0pt, 40pt}]{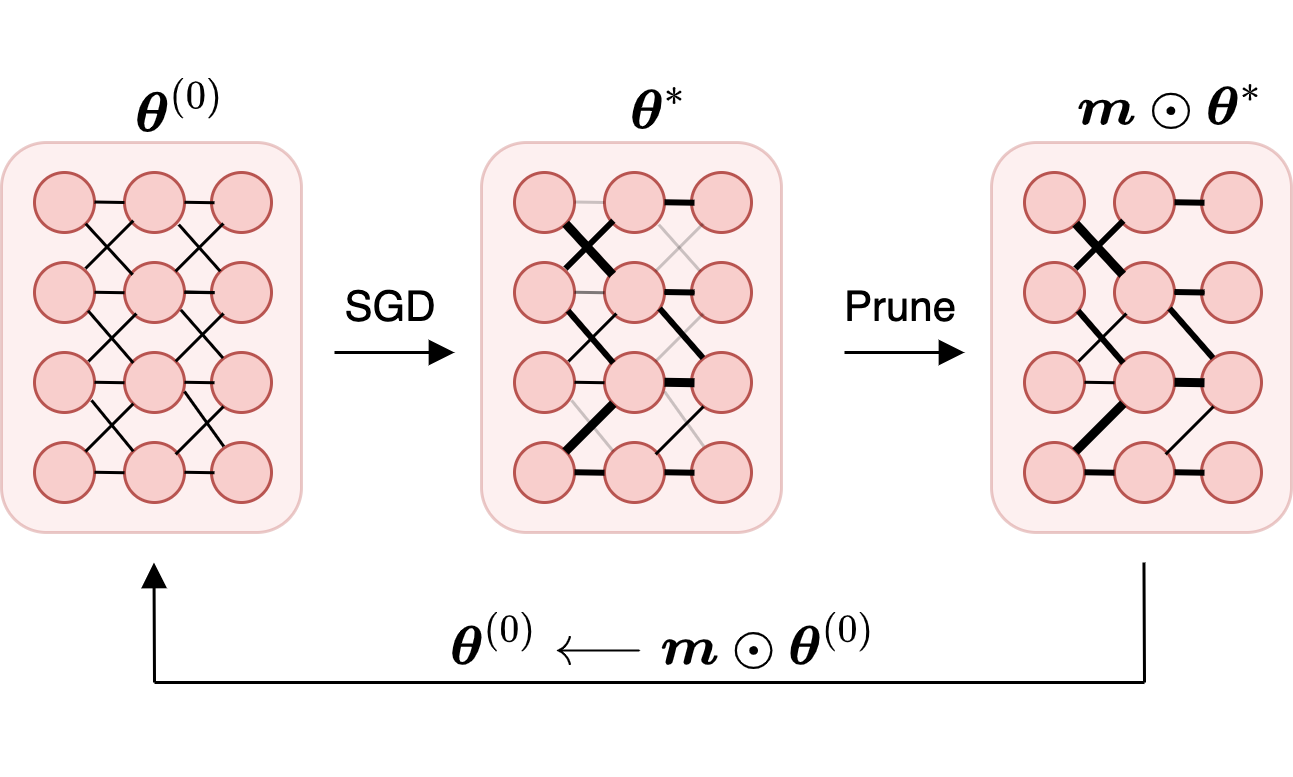}
    \caption{Illustration of the lottery ticket hypothesis and iterative magnitude-based pruning.}
    \label{fig:lottery-illustration}
\end{figure}

\section{Connection to Supervised Optimization}
\begin{figure}[t]
    \centering
    \includegraphics[width=0.99\linewidth]{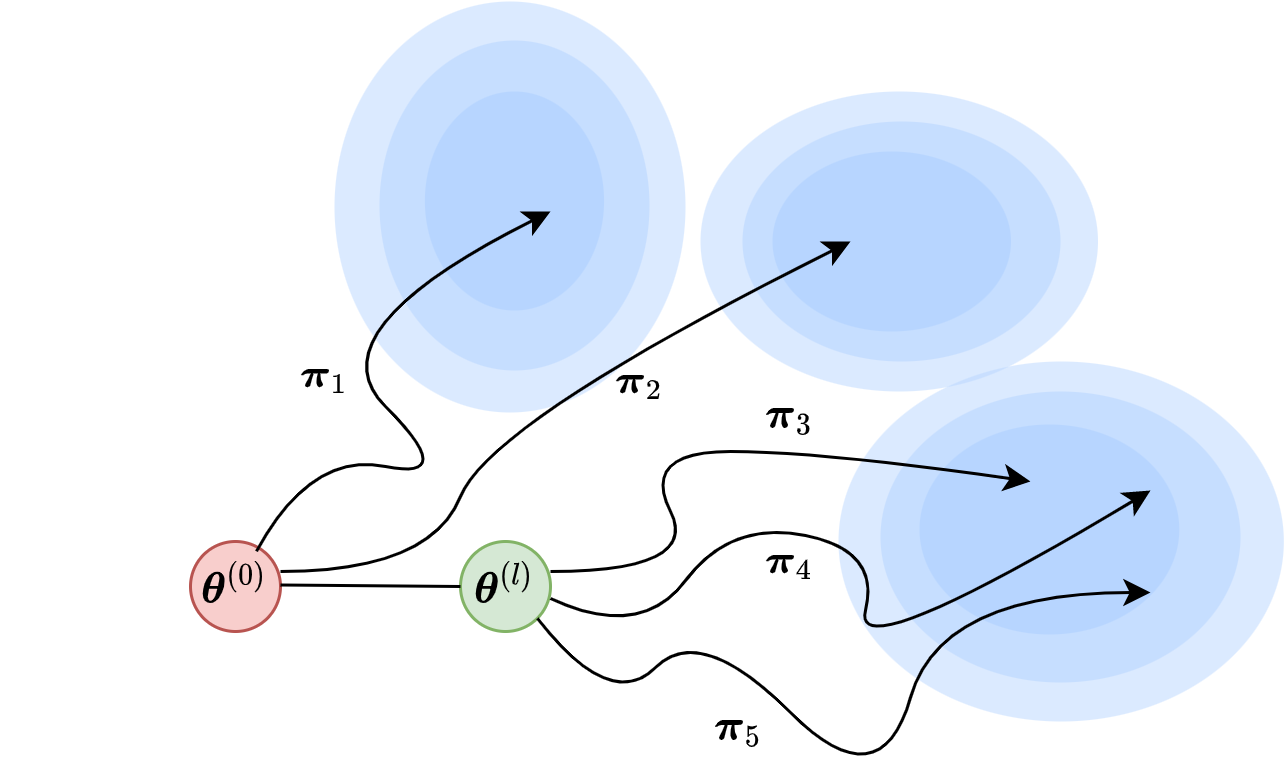}
    \caption{Illustration of stability of SGD and linear mode-connectivity. Blue contour lines indicate a basin of low test loss, $\bm{\pi}_i$ denote different batch orderings for SGD.}
    \label{fig:mode-connectivity}
\end{figure}

\paragraph{Lottery Tickets.} A way to assess the structure present in neural networks is through sparse network discovery, i.e., the \textit{lottery ticket hypothesis}. The lottery ticket hypothesis by \citet{Frankle2018TheNetworks} posits the following: Any large network possesses a sparse subnetwork that can be trained as fast and which achieves or surpasses the test error of the original network. They prove this using the power of hindsight and discover such sparse networks through the following iterative pruning strategy:
\begin{itemize}
    \item[1.] Fix an initialization $\bm{\theta}^{(0)} \sim \text{INIT}$ and train a network to convergence in a supervised fashion, leading to $\bm{\theta}^{*}$.
    \item[2.] Prune the parameters based on some criterion, leading to a binary mask $\bm{m}$ and pruned parameters $\bm{m} \odot \bm{\theta}^{*}$.
    \item[3.] Prune the initialized network $\bm{m} \odot \bm{\theta}^{(0)}$ and re-train it.
\end{itemize}
The above procedure is repeated for a fixed number of times $r$, and in every iteration, a fraction $k \in [0,1]$ of the weights is pruned, leading to an overall pruning rate of $p_r = \sum_{i=0}^{r - 1} (1 - k)^i \times k$ percentage of weights. We illustrate the algorithm in Fig.~\ref{fig:lottery-illustration}. The choice of pruning technique is flexible, but in the common variant \textit{iterative magnitude pruning (IMP)}, the globally smallest weights are pruned.  The above recipe turns out to work very well for MLPs and smaller convolutional networks, and indeed very sparse solutions can be discovered without any deterioration in terms of training time or test accuracy \citep{Frankle2018TheNetworks}. However, for more realistic architectures such as \textit{ResNets}, the picture changes and subnetworks can only be identified if the employed learning rate is low enough. Surprisingly, \citet{Frankle2019LinearHypothesis} find that subnetworks in such architectures develop very early in training and thus add the following modification to the above strategy: Instead of rewinding back to the initialization $\bm{\theta}^{(0)}$ and applying the pruning there, another checkpoint $\bm{\theta}^{(l)}$ early in training is used and $\bm{m} \odot \bm{\theta}^{(l)}$ is re-trained instead of $\bm{m} \odot \bm{\theta}^{(0)}$. \\[1mm]
\citet{Frankle2019LinearHypothesis} demonstrate that checkpoints as early as $1$ epoch can suffice to identify lottery tickets, even at standard learning rates. Interestingly, \citet{Frankle2019LinearHypothesis} further show that the point in time $l$ where lottery tickets can be found coincides with the time where SGD becomes stable to different batch orderings $\bm{\pi}$, i.e., different runs of SGD with distinct batch orderings but the same initialization $\bm{\theta}^{(l)}$ end up in the same linear basin. This property is also called linear mode connectivity; we provide an illustration in Fig.~\ref{fig:mode-connectivity}. Notice that in general, linear mode-connectivity does not hold, i.e., two SGD runs from the same initialization end up in two disconnected basins \citep{Frankle2019LinearHypothesis, Garipov2018LossDNNs}.

\begin{figure}[t]
    \centering
    \includegraphics[width=0.99\linewidth]{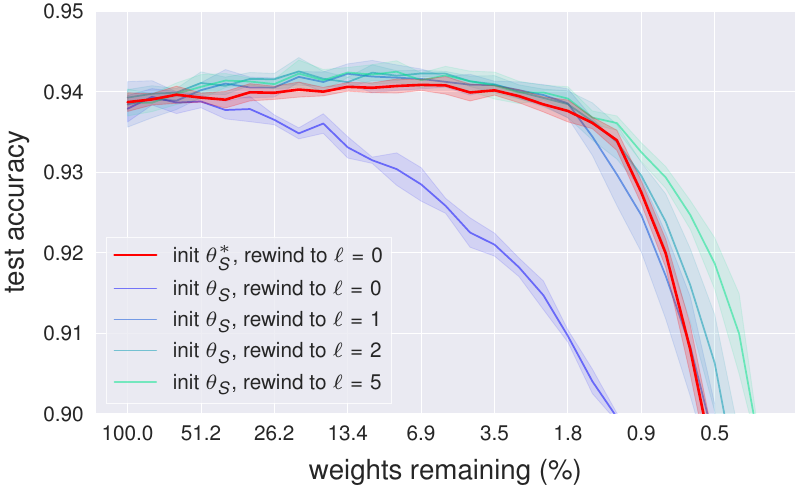}
    \caption{Test accuracy as a function of sparsity for different initialization and rewinding strategies. Fresh initializations $\bm{\theta}_S$ are not robust to IMP with rewinding to initialization ($l=0$), this only emerges with rewinding to $l \geq 1$.  Student checkpoints $\bm{\theta}_S^*$ are always robust to IMP even with rewinding to $l=0$. One epoch corresponds to 196 steps. Aggregation is done over 5 checkpoints.}
    \label{fig:imp-resnet18}
\end{figure}

\paragraph{IMP from the Student.} A natural question that emerges now is whether rewinding to a student checkpoint $\bm{\theta}_S^*$, obtained through random teacher distillation, already developed sparse structures in the form of lottery tickets. We compare the robustness of our student checkpoints $\bm{\theta}_S^*$ with random initialization at different rewinding points $\bm{\theta}^{(l)}$, closely following the setup in \citet{Frankle2019LinearHypothesis}. We display the results in Fig.~\ref{fig:imp-resnet18}, where we plot test performance on \textit{CIFAR10} as a function of the sparsity level. We use a \textit{ResNet18} and iterative magnitude pruning, reducing the network by a fraction of $0.2$ every round. We compare against rewinding to supervised checkpoints $\bm{\theta}^{(l)}$ for $l \in \{0, 1, 2, 5\}$ where $l$ is measured in number of epochs. \\[1mm]
We observe that rewinding to random initialization ($l=0$), as shown in \citet{Frankle2018TheNetworks, Frankle2019LinearHypothesis}, incurs strong losses in terms of test accuracy at all pruning levels and thus $\bm{\theta}_S$ does not constitute a lottery ticket. The distilled student $\bm{\theta}_S^{*}$, on the other hand, contains a lottery ticket, as it remains very robust to strong degrees of pruning. In fact, $\bm{\theta}_S^{*}$ shows similar behavior to the networks rewound to epoch $1$ and $2$ in supervised training. This suggests that random teacher distillation imitates some of the learning dynamics in the first epochs of supervised optimization. We stress here that  no label information was required for sparse subnetworks to develop. This aligns with results in \citep{Frankle2020TheTraining}, showing that auxiliary tasks such as rotation prediction can lead to lottery tickets. However, this is no surprise, as \citet{Anagnostidis2022TheMemorization} show that the data-informed bias of augmentations can already lead to strong forms of learning. We believe our result is more powerful since random teacher distillation relies solely on implicit regularization in SGD and does not require a task at all.

\begin{figure}[t]
    \centering
    \includegraphics[width=0.99\linewidth]{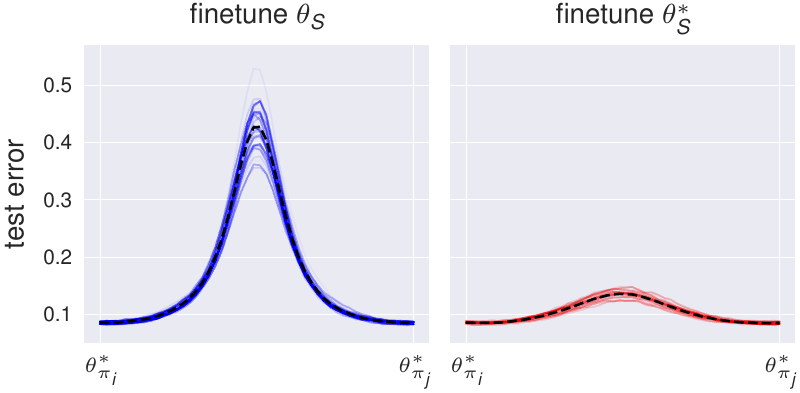}
    \caption{Test error when interpolating between networks that were trained from the same initialization. \textbf{Left:} Networks initialized at the teacher location, i.e., random initialization. \textbf{Right:} Networks initialized at the converged student $\bm{\theta}_S^{*}(0)$. Aggregation is done over $3$ initializations and $5$ different data orderings $\bm{\pi}_i$.}
    \label{fig:linear-connectivity}
\end{figure}

\paragraph{Linear Mode Connectivity.} In light of the observation regarding the stability of SGD in \citet{Frankle2019LinearHypothesis}, we verify whether a similar stability property holds for the student checkpoint $\bm{\theta}_S^{*}$. To that end, we train several runs of SGD in a supervised fashion with initialization $ \bm{\theta}_S^{*}$ on different batch orderings $\bm{\pi}_1, \dots, \bm{\pi}_b$ and study the test accuracies occurring along linear paths between different solutions $\bm{\theta}^{*}_{\bm{\pi}_i}$ for $i=1, \dots, b$, i.e.
$$\bm{\theta}_{\bm{\pi}_i \xrightarrow[]{}\bm{\pi}_j}(\gamma) := \gamma \bm{\theta}^{*}_{\bm{\pi}_i} + (1-\gamma) \bm{\theta}^{*}_{\bm{\pi}_j}.$$ 
If the test accuracy along the path does not significantly worsen, we call $\bm{\theta}^{*}_{\bm{\pi}_{\smash{i}}}$ and $\bm{\theta}^{*}_{\bm{\pi}_{\smash{j}}}$ \textit{linearly mode-connected}. We contrast the results with the interpolation curves for SGD runs started from the original, random initialization $\bm{\theta}_S$. 
We display the interpolation curves in Fig.~\ref{fig:linear-connectivity}, where we used three \textit{ResNet18} student checkpoints and finetuned each in five SGD runs with different seeds on \textit{CIFAR10}. We observe that, indeed, the resulting parameters $\bm{\theta}^{*}_{\bm{\pi}_i}$ all lie in approximately the same linear basin. However, the networks trained from the random initialization face a significantly larger barrier.
This confirms that random teacher distillation converges towards parameterizations $\bm{\theta}_S^{*}$, which are different from those at initialization $\bm{\theta}_S$. In particular, such $\bm{\theta}_S^{*}$ would only appear later in supervised optimization when SGD is already more stable to noise. Ultimately, it shows that random teacher distillation obeys similar dynamics as supervised optimization and can navigate toward linear basins of the supervised loss landscape.

\section{Discussion and Conclusion}
In this work, we examined the teacher-student setting to disentangle its implicit regularization from other very common components such as \textit{dark knowledge} in trained teachers or data augmentations in self-supervised learning. Surprisingly, students learned strong structures even from random teachers in the absence of data augmentation. We studied the quality of the students and observed that (1) probing accuracies significantly improve over the teacher, (2) features are data-dependent and transferable across tasks, and (3) student checkpoints develop sparse subnetworks at the border of linear basins without training on a supervised task. \\[1mm]
The success of teacher-student frameworks such as knowledge distillation and non-contrastive learning can thus at least partially be attributed to the regularizing nature of the learning dynamics. These label-independent dynamics allow the student to mimic the early phase of supervised training by navigating the supervised loss landscape without label information. The simple and minimal nature of our setting makes it an ideal test bed for better understanding this early phase of learning. We hope that future theoretical work can build upon our simplified framework.

\section*{Acknowledgements}
We thank Sidak Pal Singh for his valuable insights and interesting discussions on various aspects of the topic.

\bibliography{refs, mendeley}
\bibliographystyle{icml2023}

\newpage
\appendix
\onecolumn

\section{The Algorithm}

Distillation from a random teacher has two important details. The outputs are very high-dimensional, $2^{16}$-d. And a special component, the \emph{l2-bottleneck}, is hidden in the architecture of the projection head just before the softmax. It linearly maps a feature vector to a low-dimensional space, normalizes it, and computes the dot product with a normalized weight matrix, i.e.
$$ x \rightarrow 
\tilde{V}^T 
\frac{W^T x + b}{||W^T x + b||_2} 
\;\text{ with } 
||\tilde{V}_{:,i}||_2 = 1 $$ 
for 
$x \in \mathbb{R}^{n}$, 
$W \in \mathbb{R}^{n \times k}$, 
$b \in \mathbb{R}^{k}$, 
$\tilde{V} \in \mathbb{R}^{k \times m}$. This architecture is heavily inspired by DINO~\cite{Caron2021EmergingTransformers}. Let us summarize the method in pseudo-code:

\begin{lstlisting}[language=Python, mathescape=true]
encoder, head, wn_layer = ResNet(512), MLP(2048,2048,256), Linear($2^{16}$)

student = initialize(encoder, head, wn_layer)
teacher = copy(student) # initialize with same parameters
for x, y in repeat(data, n_epochs):
    # apply weight-normalization
    normalized_weight_t = normalize(teacher.wn_layer.weight)
    normalized_weight_s = normalize(student.wn_layer.weight)

    # prepare target
    x_t = teacher.head(teacher.encoder(x))
    x_t = normalize(x_t)
    x_t = dot(normalized_weight_t, x_t)
    target = softmax(x_t)
    
    # prepare prediction
    x_s = student.head(student.encoder(x))
    x_s = normalize(x_s)
    x_s = dot(normalized_weight_s, x_s)
    prediction = softmax(x_s)
    
    # compute loss, backpropagate and update
    loss = sum(target * -log(prediction)) # cross-entropy
    loss.backward()
    optimizer.step(student) # update only student
\end{lstlisting}

\noindent

\begin{figure}[h!]
    \centering
    \vspace{-0.5em}
    \includegraphics[width=0.75\textwidth]{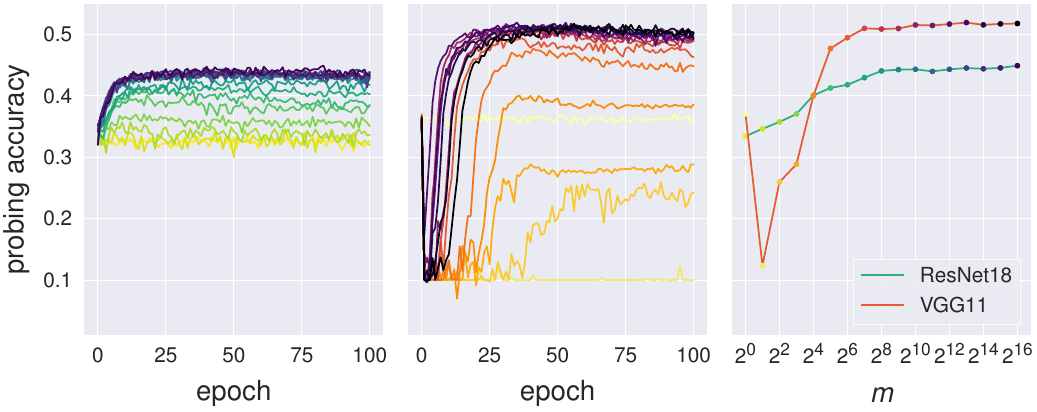}
    \vspace{-1em}
    \caption{Comparing different output dimensions $m$ of the projection head. Large $m=2^{16}$ are not crucial for feature learning, but there is phase transition at the bottleneck dimension $m=2^8=256$ Linear probing on \textit{CIFAR10}.
    \textbf{Left:} \textit{ResNet18} (red). \textbf{Right:} \textit{VGG11} (green).}
    \label{fig:init-probe-outdim}
    \vspace{-1.5em}
\end{figure}

\newpage
\section{Ablating the Projector}
\label{app:projector}
\subsection{Ablating Normalization Layers}
\begin{figure}[ht]
    \centering
    \vspace{-0.5em}
    \includegraphics[width=0.75\textwidth]{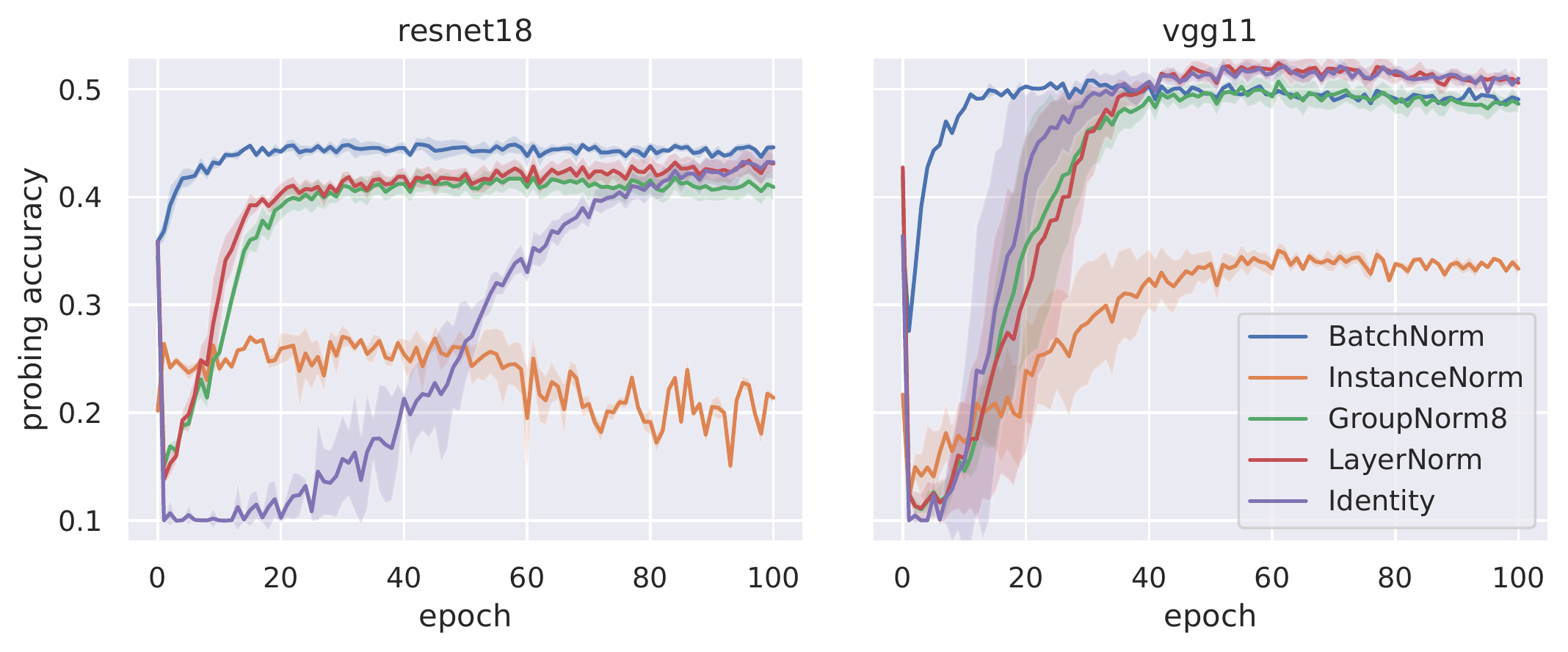}
    \vspace{-1em}
    \caption{Comparing different types of normalization layers on \textit{CIFAR10}. \textbf{Left}: \textit{ResNet18}. \textbf{Right}: \textit{VGG11}.}
    \label{fig:init-probe-normlayer}
    \vspace{-1.5em}
\end{figure}

If the teacher is used in evaluation mode, then one possible source of asymmetry is introduced by batch normalization layers. But is the effect caused by this batch-dependent signal? Or does the batch dependency amplify the mechanism? In Fig.~\ref{fig:init-probe-normlayer} we compare different types of normalization layers and no normalization (Identity). We observe that although BN stabilizes training, the effect also occurs with batch-independent normalization. Further, networks without normalization reach similar performance but take longer to converge.

\subsection{Ablating the L2-Bottleneck}
\begin{figure}[ht]
    \centering
    \vspace{-0.5em}
    \includegraphics[width=0.75\textwidth]{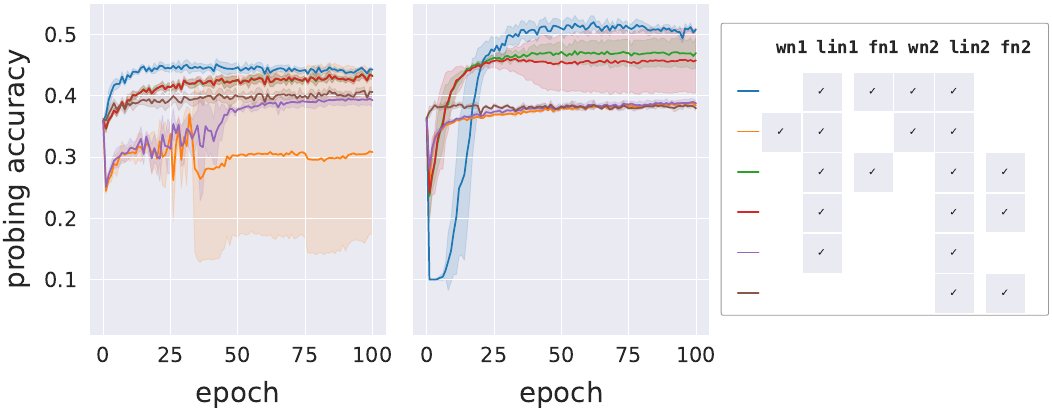}
    \vspace{-1em}
    \caption{Ablating components of the \emph{l2-bottleneck} on \textit{CIFAR10}. \textbf{Left}: \textit{ResNet18}. \textbf{Right}: \textit{VGG11}.}
    \label{fig:init-probe-l2bot}
    \vspace{-1.5em}
\end{figure}

The \textit{l2-Bottleneck} is a complex layer with many unexplained design choices. We compare different combinations of weight-normalization (wn), linear layer (lin), and feature normalization (fn) for the first and second part of the bottleneck in Figures~\ref{fig:init-probe-l2bot} for a~\textit{ResNet18} and a~\textit{VGG11} respectively. While the default setup is clearly the most performant, removing feature normalization is more destructive than removing weight normalization. In particular, only one linear layer followed by a feature normalization still exhibits a similar trend and does not break down.

\newpage
\section{Additional Results}
\label{app:more-stuff}
We present additional experimental results that serve to better understand the regularization properties of self-distillation with random teachers.

\subsection{$K$-NN probing}
\label{app:more-probe}
A different probing choice, instead of learning a linear layer on top of the extracted embeddings, is to perform $K$-NN classification on the features. We apply $K$-nearest-neighbour classification with the number of neighbors set to $K=20$, as commonly done in practice. As in Table~\ref{tab:distill} in the main text, we present results under $K$-NN evaluation in Table~\ref{tab:distill-knn}. Also, as in Table~\ref{tab:distill_transfer}, we evaluate using $K$-NN probing the transferability of the learned embeddings from \textit{TinyImageNet} in Table~\ref{tab:distill_transfer-knn}.

\begin{table}[h!]
\vskip 0.15in
\begin{center}
\begin{small}
\begin{sc}
\begin{tabular}{lcccc}
\toprule
Dataset & Model & Teacher & Student & Input \\
\midrule
\multirow{ 2}{*}{ \textit{CIFAR10}}  
    & \textit{ResNet18} & $37.65$ & $\mathbf{44.67}$ & \multirow{2}*{$33.61$} \\[1mm]
    & \textit{VGG11} & $44.92$ & $\mathbf{51.32}$ &   \\
\midrule
\multirow{ 2}{*}{ \textit{CIFAR100}}  
    & \textit{ResNet18} & $13.77$ & $\mathbf{20.22}$ & \multirow{2}*{$14.87$} \\[1mm]
    & \textit{VGG11} & $18.10$ & $\mathbf{23.53}$ & \\
\midrule
\multirow{ 2}{*}{ \textit{STL10}}  
    & \textit{ResNet18} & $31.71$ & $\mathbf{37.41}$ & \multirow{2}*{$28.94$} \\[1mm]
    & \textit{VGG11} & $36.92$ & $\mathbf{43.58}$ & \\
\midrule
\multirow{ 2}{*}{ \textit{TinyImageNet}} 
    & \textit{ResNet18} & $4.59$ & $\mathbf{7.11}$ & \multirow{2}*{$3.44$} \\[1mm]
    & \textit{VGG11} & $5.98$ & $\mathbf{9.23}$ & \\
\bottomrule
\end{tabular}
\end{sc}
\end{small}
\end{center}
\vskip -0.11in
\caption{$K$-NN probing accuracies (in percentage) of the representations for various datasets for teacher, student, and raw pixel inputs.}
\label{tab:distill-knn}
\end{table}

\begin{table}[h!]
\vskip 0.15in
\begin{center}
\begin{small}
\begin{sc}
\begin{tabular}{lccc}
\toprule
Dataset & Model & Teacher & Student \\
\midrule
\multirow{ 2}{*}{ \textit{CIFAR10}}  
    & \textit{ResNet18} & $37.65$ & $\mathbf{44.45}$  \\[1mm]
    & \textit{VGG11} & $44.92$ & $\mathbf{51.48}$ \\
\midrule
\multirow{ 2}{*}{ \textit{CIFAR100}}  
    & \textit{ResNet18} & $13.77$ & $\mathbf{19.48}$ \\[1mm]
    & \textit{VGG11} & $18.10$ & $\mathbf{23.95}$  \\
\midrule
\multirow{ 2}{*}{ \textit{STL10}}  
    & \textit{ResNet18} & $31.71$ & $\mathbf{38.86}$ \\[1mm]
    & \textit{VGG11} & $36.92$ & $\mathbf{42.26}$  \\
\bottomrule
\end{tabular}
\end{sc}
\end{small}
\end{center}
\vskip -0.11in
\caption{$K$-NN probing accuracies (in percentage) of the representations for various datasets for teacher and student when transferred from \textit{TinyImageNet}.}
\label{tab:distill_transfer-knn}
\end{table}

\newpage
\subsection{Architectures} \label{app:more-architectures}
For our experiments in the main text, we used the very common \textit{VGG11} and \textit{ResNet18} architectures. Here, we report results for different types of architectures to provide a better picture of the relevance of architectural inductive biases. In particular, we compare with the \textit{Vision Transformer (ViT)} \cite{dosovitskiy2020image} (patch size $8$ for $32\times32$ images of \textit{CIFAR10}) and find that the effect of representation learning is still present, albeit less pronounced. More generally, we observe that with less inductive bias, the linear probing accuracy diminishes but never breaks down.

\begin{table}[!h]
\vskip 0.15in
\begin{center}
\begin{small}
\begin{scshape}
\begin{tabular}{lrccc}
\toprule
Model & \#params & Teacher & Student  \\

\midrule
None (Input) 
    & $0$ & $39.02$ & $39.02$ \\

\midrule
\textit{VGG11} 
    & $9'220'480$ & $36.55$ & $\mathbf{51.98}$ \\[1mm]
\textit{VGG13} 
    & $9'404'992$ & $34.73$ & $\mathbf{49.26}$ \\[1mm]
\textit{VGG16} 
    & $14'714'688$ & $33.08$ & $\mathbf{46.35}$ \\[1mm]
\textit{VGG19} 
    & $20'024'384$ & $30.84$ & $\mathbf{43.90}$ \\

\midrule
\textit{ResNet20*} 
    & $271'824$ & $28.68$ & $\mathbf{36.62}$ \\[1mm]
\textit{ResNet56*} 
    & $855'120$ & $14.05$ & $\mathbf{27.92}$ \\[1mm]
\textit{ResNet18} 
    & $11'168'832$ & $35.50$ & $\mathbf{46.02}$ \\[1mm]
\textit{ResNet34} 
    & $21'276'992$ & $28.18$ & $\mathbf{41.04}$ \\[1mm]
\textit{ResNet50} 
    & $23'500'352$ & $19.69$ & $\mathbf{27.53}$ \\
    
\midrule
\textit{ViT-Tiny} 
    & $594'048$ & $32.93$ & $\mathbf{35.76}$ \\[1mm]
\textit{ViT-Small} 
    & $2'072'832$ & $38.57$ & $\mathbf{41.68}$ \\[1mm]
\textit{ViT-Medium} 
    & $3'550'208$ & $41.09$ & $\mathbf{43.13}$ \\[1mm]
\textit{ViT-Base} 
    & $7'684'608$ & $41.71$ & $\mathbf{44.38}$  \\

\bottomrule
\end{tabular}
\end{scshape}
\end{small}
\end{center}
\caption{Linear probing accuracies (in percentage) of the representations for various architectures for teacher, student, and flattened inputs on \textit{CIFAR10}. \textit{ResNet20*} and \textit{ResNet56*} are the smaller CIFAR-variants from \citet{he2015resnet}. The students outperform their teachers in all cases.}
\label{tab:more-arch}
\end{table}

\newpage
\subsection{Loss landscapes}
\label{app:more-loss}

\begin{figure}[ht]
    \centering
    \includegraphics[width=0.5\textwidth]{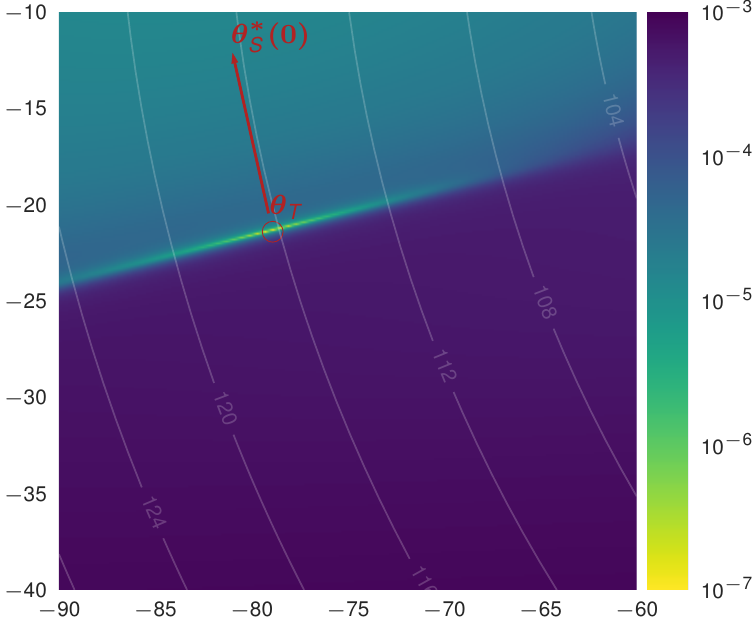}
    \caption{Higher resolution crop of the global optimum around the teacher.}
    \label{fig:losslandscape_ortho_zoom}
\end{figure}

The parameter plane visualized in Fig.~\ref{fig:losslandscape} is defined by interpolation between three parameterizations, thus, distances and angles are not preserved. In the following Fig.~\ref{fig:losslandscape_ortho}, we orthogonalize the basis of the parameter plane to achieve a distance and angle-preserving visualization. We note that both converged solutions of the students $\bm{\theta}_S^{*}(0)$ and $\bm{\theta}_S^{*}(1)$ stay comparably close to their initializations.
Further, we provide a zoomed crop of the asymmetric valley around the teacher $\bm{\theta}_{S_T}$ in Fig.~\ref{fig:losslandscape_ortho_zoom}.
\begin{figure}[ht]
    \centering
    \includegraphics[width=0.9\textwidth]{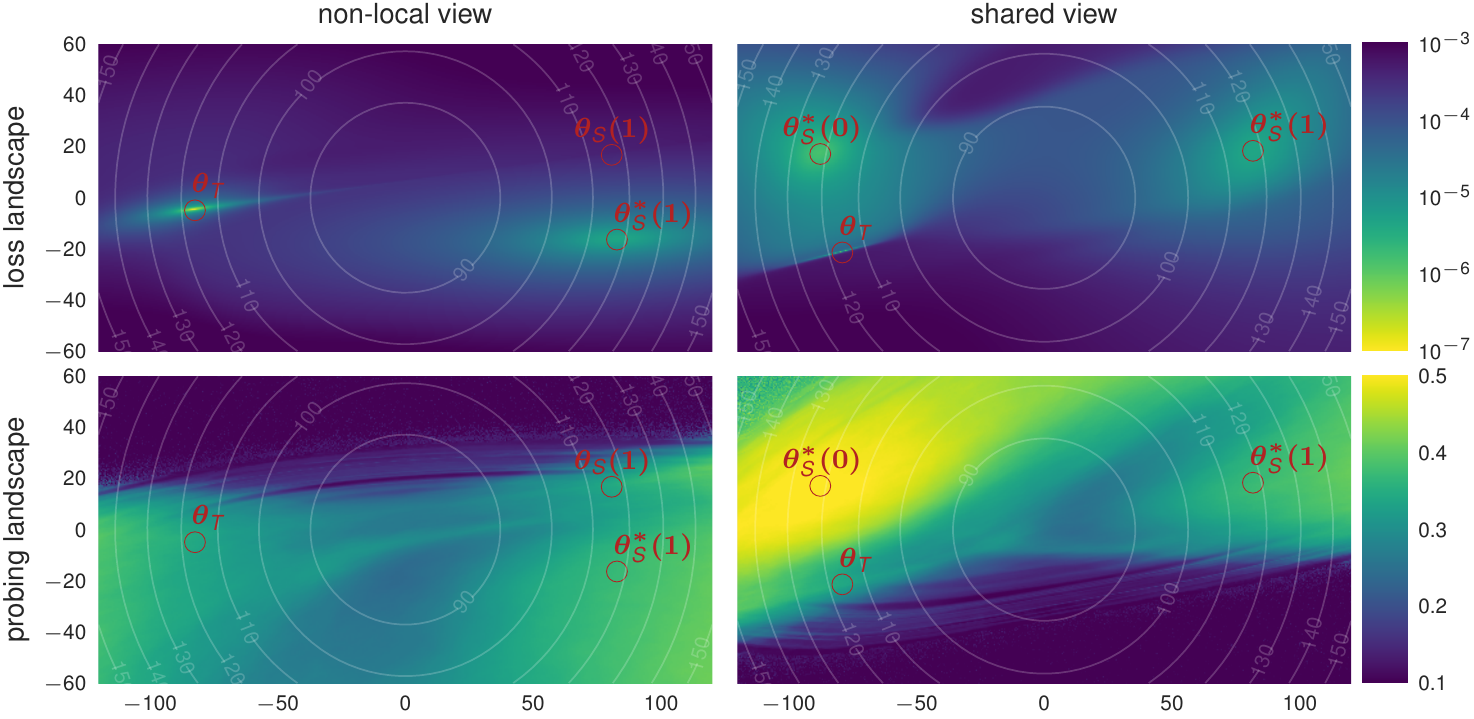}
    \caption{Orthogonal projection of the loss landscape in the parameter plane.}
    \label{fig:losslandscape_ortho}
\end{figure}

\newpage
The same visualization technique allows plotting the KL divergence between embeddings produced by the teacher and other parametrization in the plane. While in Fig,\ref{fig:losslandscape_ortho}, the basin of the local solution matches with the area of increased probing accuracy, such a correlation is not visible if one only considers the encoder. 
\begin{figure}[ht]
    \centering
    \includegraphics[width=0.9\textwidth]{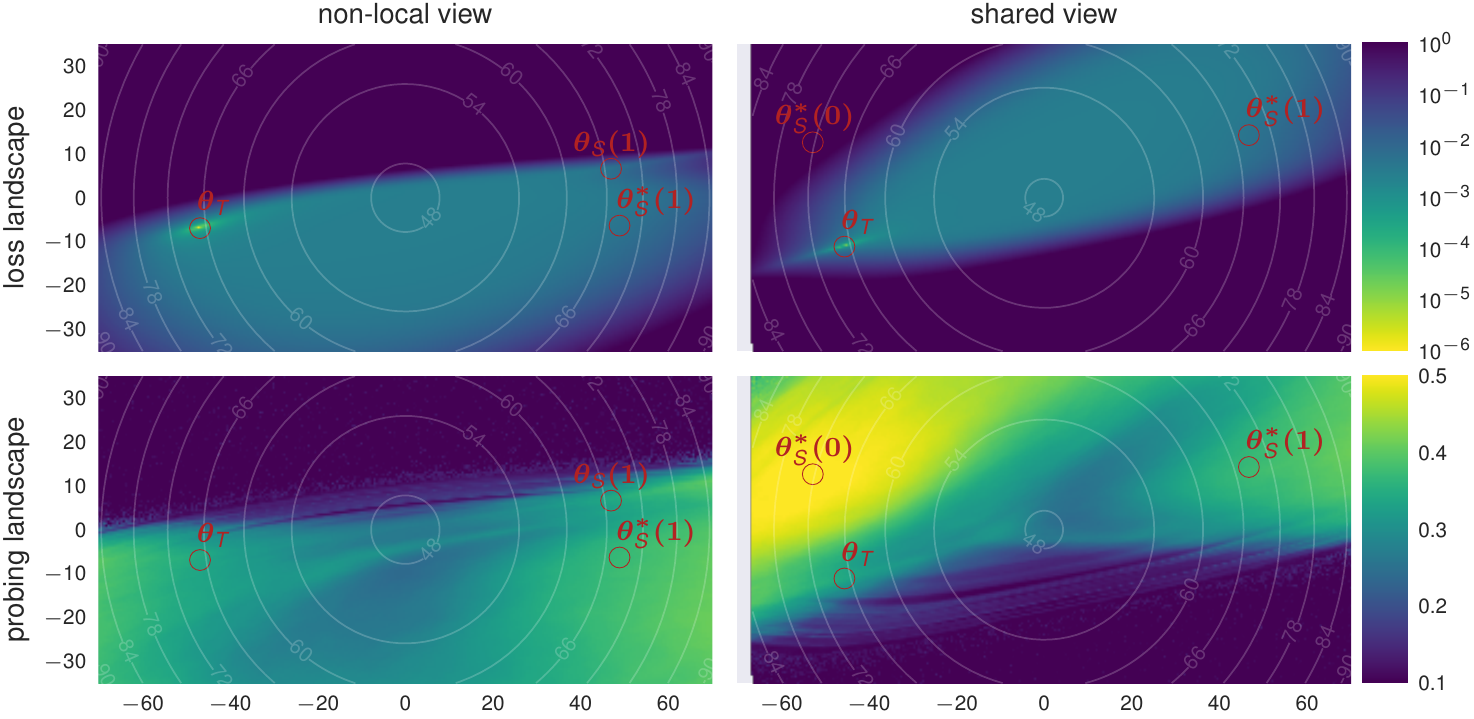}
    \caption{Orthogonal projection of the embedding KL divergence landscape in the parameter plane.}
    \label{fig:losslandscape_enc_ortho}
\end{figure}
\begin{figure}[ht]
    \centering
    \includegraphics[width=0.6\textwidth]{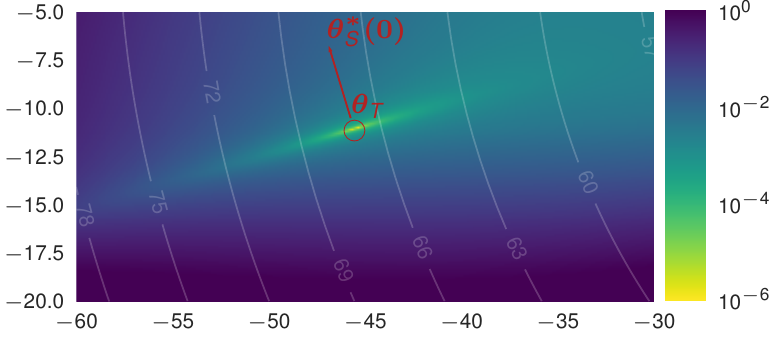}
    \caption{Higher resolution crop of the global optimum around the teacher.}
    \label{fig:losslandscape_enc_ortho_zoom}
\end{figure}

\newpage
\section{Optimization Metrics} \label{app:metrics}
\begin{figure}[ht]
    \centering
    \includegraphics[width=0.65\linewidth]{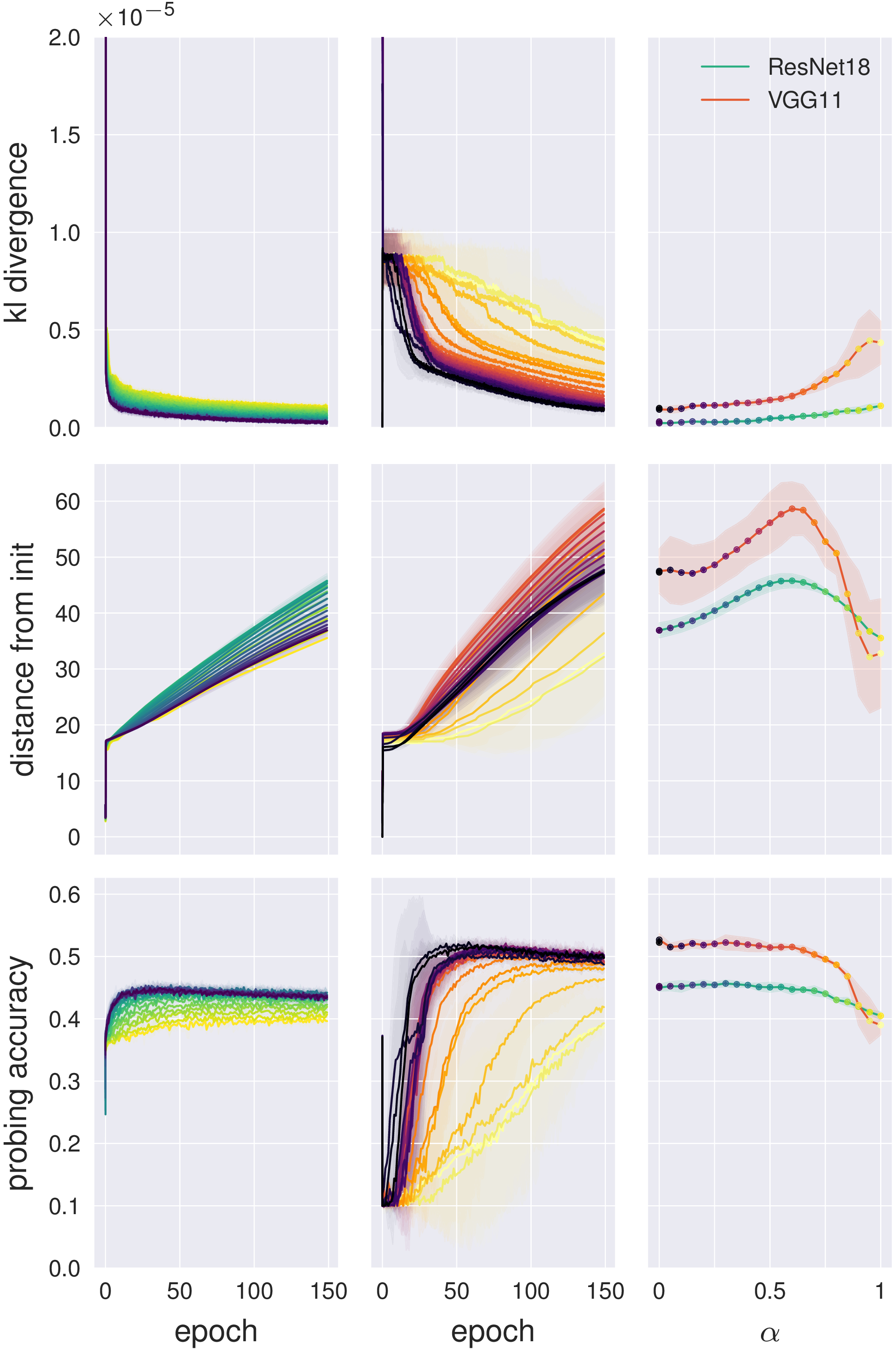}
    \caption{Optimization metrics for locality parameter $\alpha$ on \textit{CIFAR10}.    \textbf{Left:} \textit{ResNet18}. \textbf{Middle:} \textit{VGG11}. \textbf{Right:} Summary.}
    \label{fig:interpolate_traj}
\end{figure}

To convince ourselves that independently initialized students ($\alpha=1$) are more difficult to optimize, we provide an overview of the KL-Divergence and distance from initialization for all $\alpha \in [0,1]$ in Fig.~\ref{fig:interpolate_traj}. We observe that, indeed, for students initialized far away from their teacher, the loss cannot be reduced as efficiently. This coincides with worse probing performance. Note, however, that even the students with $\alpha=1$ are able to outperform their teachers.

\newpage
\subsection{Restarting} \label{app:restart}
\begin{figure}[ht]
    \centering
    \includegraphics[width=0.64\linewidth]{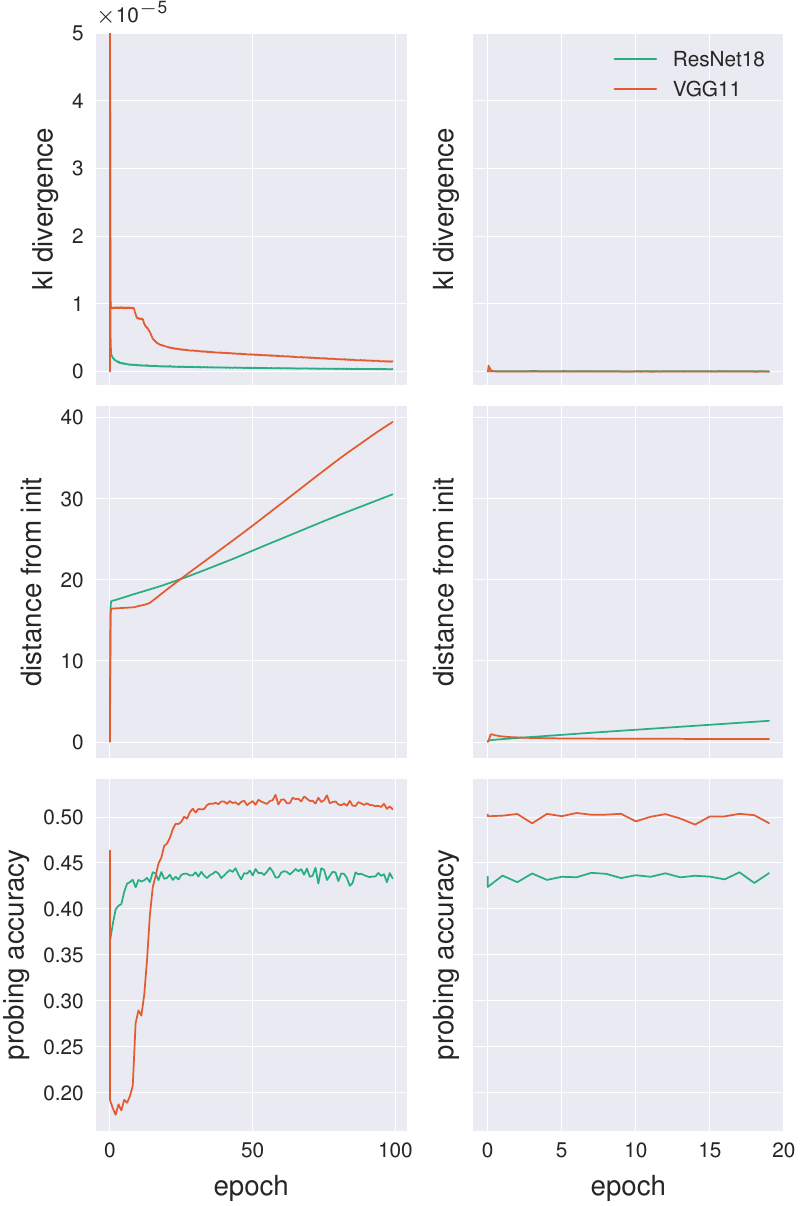}
    \caption{Restarting random teacher distillation on \textit{CIFAR10} with \textit{ResNet18} and \textit{VGG11}. \textbf{Left:} First round of distillation. \textbf{Right:} Second round of distillation}
    \label{fig:restart}
\end{figure}

An evident idea would be to restart the random teacher distillation procedure in some way or another. We considered several approaches, such as reintroducing the exponential moving average of the teacher, but were not successful. In Fig.~\ref{fig:restart}, we show the most straightforward approach, where the student is reused as a new teacher, and a second round of distillation is performed. The gradient dynamics around the restarted student seem much more stable, and the optimization procedure does not even begin.

\newpage
\section{Experimental Details}
\label{app:details}
Our main goal is to demystify the properties of distillation in a simplistic setting, removing a series of `tricks' used in practice. For clarity reasons, we here present a comprehensive comparison with the popular framework of DINO~\citep{Caron2021EmergingTransformers}.

\subsection{Architecture} \label{lab:architecture}
\begin{tabularx}{\textwidth}{l|l}
    Configuration & \\
    \hline
    Encoder 
        & ResNet18\&VGG1 from torchvision, without fc or classification layers (embedding $\in \mathbb{R}^{512}$)  \\
        & (ResNet18 adjusted stem for CIFAR: conv from 7x7 to 3x3, remove maxpool) \\
    Projection Head 
        & 3-Layer MLP: $512 \rightarrow 2048 \rightarrow 2048 \rightarrow \text{l2-bottleneck}(256) \rightarrow 2^{16}$ \\
        & (GELU activation, no batchnorms, init: trunc\_normal with $\sigma=0.02$, biases=0) \\
    L2-Bottleneck(in, mid, out)
        & for $x \in \mathbb{R}^{in}$, $W \in \mathbb{R}^{in \times mid}$, $b \in \mathbb{R}^{mid}$, $\tilde{V} \in \mathbb{R}^{mid \times out}$ \\
        & 1. linear to bottleneck: $z = W^T x + b \in \mathbb{R}^{mid}$ \\
        & 2. feature normalization: $\tilde{z} = z/||z||_2$\\
        & 3. weightnormalized linear: $y = \tilde{V}^T \tilde{z} \in \mathbb{R}^{out}$, with $||\tilde{V}_{:,i}||_2 = 1$ \\
        & \quad $ \Rightarrow f_{\tilde{V},W}(x) = 
        \tilde{V}^T \frac{W^T x + b}{||W^T x + b||_2} 
        \;\text{ with } ||\tilde{V}_{:,i}||_2 = 1  $
\end{tabularx}

\subsection{Data}
\begin{tabularx}{\textwidth}{l|l|l}
    Configuration & DINO default & Random Teacher \\
    \hline
    Augmentations 
        & Multicrop ($2\times224^2 + 10\times96^2$) + SimCLR-like
        & \textbf{None}  ($\mathbf{1\times32^2}$)
    \\ Training batchsize 
        & 64 per GPU
        & 256 
    \\ Evaluation batchsize 
        & 128 per GPU
        & 256
\end{tabularx}

\subsection{DINO Hyperparameters}
\begin{tabularx}{\textwidth}{l|l|l}
    Configuration & DINO default & Random Teacher\\
    \hline
    Teacher update 
        & ema with momentum $0.996\stackrel{cos}{\rightarrow}1$
        & no updates
    \\ Teacher BN update 
        & BN in train mode
        & BN in eval mode
    \\ Teacher centering 
        & track statistics with momentum 0.9
        & not applied
    \\ Teacher sharpening 
        & temperature 0.04 (paper: $0.04\stackrel{lin}{\rightarrow}0.07$)
        & temperature 1
    \\ Student sharpening 
        & temperature 0.1
        & temperature 1
    \\ Loss function 
        & opposite-crop cross-entropy
        & single-crop cross-entropy
\end{tabularx}

\subsection{Random Teacher Training}
\begin{tabularx}{\textwidth}{l|l|l}
    Configuration & DINO default & Random Teacher  \\
    \hline
    Optimizer 
        & AdamW 
        & AdamW 
    \\ Learning rate 
        & $0 \stackrel{lin}{\rightarrow} 0.0005 \stackrel{cos}{\rightarrow} \texttt{1e-6} $ schedule 
        & 0.001 (torch default)
    \\ Weight decay 
        & $0.04 \stackrel{lin}{\rightarrow} 0.4$ schedule 
        & not applied
    \\ Gradient Clipping 
        & to norm 3 
        & not applied
    \\ Freezing of last layer 
        & during first epoch 
        & not applied
\end{tabularx}

\subsection{IMP Training}
\begin{tabularx}{\textwidth}{l|l|l}
    Configuration & Lottery Ticket Hypothesis \cite{Frankle2020TheTraining} & Random Teacher  \\
    \hline
    Training Epochs 
        & 160
        & 160
    \\ Optimizer 
        & SGD (momentum 0.9)
        & SGD (momentum 0.9)
    \\ Learning rate 
        & MultiStep: $0.1  \stackrel{80 \text{ epochs}}{\rightarrow} 0.01 \stackrel{40 \text{ epochs}}{\rightarrow} 0.001$
        & MultiStep: $0.1  \stackrel{80 \text{ epochs}}{\rightarrow} 0.01 \stackrel{40 \text{ epochs}}{\rightarrow} 0.001$
    \\ Weight decay 
        & 0.0001
        & 0.0001 
    \\ Augmentations 
        & Random horizontal flip \& padded crop (4px)
        & Random horizontal flip \& padded crop (4px)
\end{tabularx}

\end{document}